%% file: Paper-0428.tex
\documentclass[runningheads]{llncs}
\usepackage[T1]{fontenc}
\usepackage{graphicx,verbatim}
\usepackage{url}
\usepackage{graphicx}
\usepackage{booktabs}
\usepackage{multirow}
\usepackage{colortbl}
\usepackage{xcolor}
\usepackage{amsmath,amssymb,graphicx}
\usepackage{bm}
\usepackage{bbm}
\usepackage{graphicx, wrapfig}
\definecolor{customblue}{RGB}{75,75,225}
\usepackage{makecell}

\begin{document}
\title{Pathology-Aware Adaptive Watermarking\\for Text-Driven Medical Image Synthesis}
\titlerunning{Pathology-Aware Watermarking for Text-Driven Medical Image Synthesis}

\def\thefootnote{$*$}\footnotetext{Equal contribution\quad$^{\dagger}$ Corresponding Author\quad $^{\ddagger}$ Work done prior to joining Amazon}

\author{Chanyoung Kim$^{*}$\inst{1}\quad  
Dayun Ju$^{*}$\inst{1}\quad  
Jinyeong Kim\inst{1}\quad  
Woojung Han\inst{1}\\  
Roberto Alcover-Couso\inst{2}\quad  
Seong Jae Hwang$^{\dagger}$\inst{1}}  

\authorrunning{Kim et al.}
\institute{Department of Artificial Intelligence, Yonsei University, Republic of Korea\and Amazon, Madrid, Spain$^{\ddagger}$\\
\email{\tt\small \{chanyoung, juda0707, jinyeong1324, dnwjddl, seongjae\}@yonsei.ac.kr, ralcover@amazon.com}
}

\maketitle
\input{tex/abstract}
\input{tex/introduction}
\input{tex/method}
\input{tex/experiments}
\input{tex/conclusion}


\subsubsection{\ackname} 
This work was supported in part by the IITP RS-2024-00457882 (AI Research Hub Project), IITP 2020-II201361, NRF RS-2024-00345806, NRF RS-2023-00219019, and NRF RS-2023-002620.

\bibliographystyle{splncs04}
\bibliography{Paper-0428}
\end{document}

%% file: tex/abstract.tex
\begin{abstract}
As recent text-conditioned diffusion models have enabled the generation of high-quality images, concerns over their potential misuse have also grown.
This issue is critical in the medical domain, where text-conditioned generated medical images could enable insurance fraud or falsified records, highlighting the urgent need for reliable safeguards against unethical use.
While watermarking techniques have emerged as a promising solution in general image domains, their direct application to medical imaging presents significant challenges.
A key challenge is preserving fine-grained disease manifestations, as even minor distortions from a watermark may lead to clinical misinterpretation, which compromises diagnostic integrity.
To overcome this gap, we present MedSign, a deep learning-based watermarking framework specifically designed for text-to-medical image synthesis, which preserves pathologically significant regions by adaptively adjusting watermark strength.
Specifically, we generate a pathology localization map using cross-attention between medical text tokens and the diffusion denoising network, aggregating token-wise attention across layers, heads, and time steps.
Leveraging this map, we optimize the LDM decoder to incorporate watermarking during image synthesis, ensuring cohesive integration while minimizing interference in diagnostically critical regions.
Experimental results show that our MedSign preserves diagnostic integrity while ensuring watermark robustness, achieving state-of-the-art performance in image quality and detection accuracy on MIMIC-CXR and OIA-ODIR datasets.

\keywords{Image Watermarking  \and Diffusion Models \and Cross Attention}

\end{abstract}

%% file: tex/introduction.tex
\section{Introduction}
Recent advancements in diffusion probabilistic models~\cite{han2025spatial,ho2020denoising,glide,rombach2022high} have enabled the generation of highly realistic synthetic images, benefiting applications such as content creation~\cite{ramesh2022hierarchical} and data augmentation~\cite{trabucco2024effective}.
However, as generative models become more powerful, concerns over their potential misuse have grown, leading to the introduction of regulatory measures such as the EU AI Act~\cite{EU_AI_Act_2024}, Korean AI Basic Act~\cite{Korea_Law_2024}, and Chinese AI governance rules~\cite{China_AI_Plan_2023}. 

A widely accepted solution to mitigate such risks is watermarking, which embeds imperceptible yet robust signals into generated content.
Traditional signal-based methods~\cite{al2007combined} are fragile to transformations, prompting the rise of deep learning-based approaches~\cite{fernandez2023stable,jang2024waterf,sander2025watermark,zhang2024editguard} that embed in image or feature space. Recent semantic-aware strategies~\cite{ci2024ringid,wen2023tree} enhance durability by manipulating latent features, though they may affect image structure. As a result, general-purpose deep watermarking methods continue to gain traction for balancing robustness and fidelity.

This need becomes even more critical in the medical domain, where text-driven image generation~\cite{Bluethgen2024,2024cxrl,lee2024llmcxr,medghalchi2024meddap,Yellapragada_2024_WACV} enables the synthesis of highly realistic images from clinical descriptions. While effective, such capabilities raise concerns, as misuse could directly impact clinical decisions.
Any manipulation of medical images can compromise patient care, violate confidentiality, and lead to serious real-world consequences~\cite{mirsky2019ct}.
For instance, fabricated medical images could be exploited for insurance fraud by falsely indicating the presence of a disease. 
Moreover, synthetic medical images could be altered to create misleading medical records, falsely attributing medical conditions to individuals, potentially leading to legal ramifications and a breach of public trust. 
Hence, there is an urgent need for watermarking methods that can ensure the authenticity and integrity of \textit{text-driven synthetic medical images}, safeguarding against potential misuse.

However, unlike general images, medical images exhibit distinct characteristics that make watermarking particularly challenging. 
A critical requirement is \textit{the preservation of fine-grained anatomical details} (e.g., detailed tissue texture), as even subtle modifications can affect clinical interpretations~\cite{siracusano2023effective}.  
For instance, in chest X-ray (CXR) images, minor alterations in opacity patterns may mimic or obscure signs of pulmonary edema, potentially leading to misdiagnosis. 
Therefore, a watermarking method for medical images must embed a watermark message while maintaining diagnostic integrity and ensuring security.

To address these challenges, we introduce MedSign, a pioneering deep watermarking framework for text-to-medical image synthesis. 
To preserve diagnostic integrity, we leverage the diffusion model’s internal cross-attention representation~\cite{Marcos-Manchon_2024_CVPR,vaswani2017attention} to identify critical regions, as these attention maps highlight clinically significant features.
Leveraging this pathology-aware guidance, we fine-tune the variational autoencoder (VAE) decoder within the latent diffusion model (LDM)~\cite{rombach2022high} to regulate watermark placement, ensuring that embedding occurs only in non-critical areas while preserving diagnostically relevant structures.
This enables MedSign to achieve a balance between robust watermarking and clinical utility, securing generated medical images without compromising diagnostic reliability.
Notably, MedSign inherently integrates watermarking into the image synthesis process without modifying the LDM's architecture or requiring additional post-processing of the generated image.
Experiments on CXRs~\cite{johnson2019mimic} and fundus images~\cite{li2021benchmark} confirm that MedSign effectively directs the watermark to non-critical regions while maintaining diagnostic fidelity.
Furthermore, our MedSign outperforms watermarking techniques designed for general images, achieving superior image quality and watermarking performance in the medical domain.

%% file: tex/method.tex
\section{Methods}
\begin{figure}[t!]
  \centering
  \includegraphics[width=\linewidth]{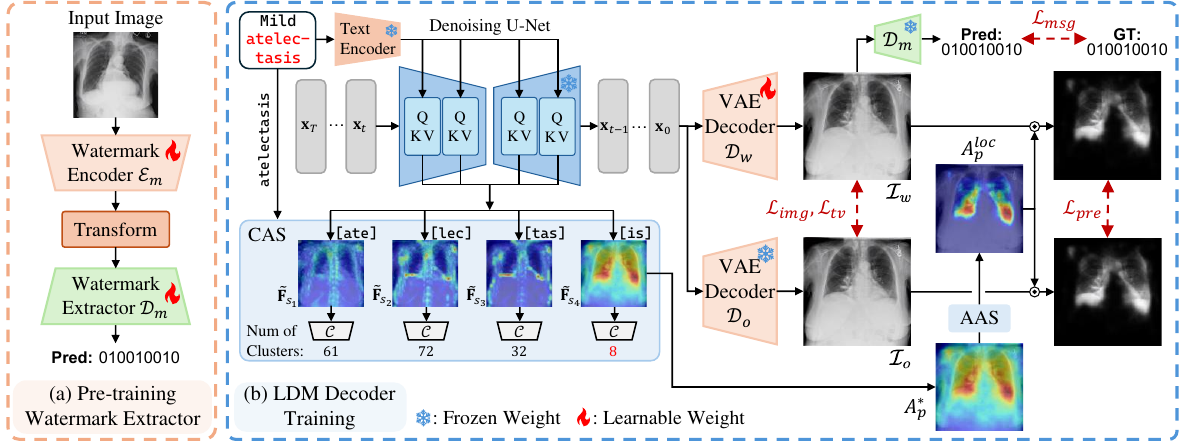}
  \caption{
  Detailed illustration of our pipeline.
  (a) We first train the watermark encoder \(\mathcal{E}_m\) and extractor \(\mathcal{D}_m\) to embed and extract watermarks from transformed images. 
  (b) We then train the LDM’s VAE decoder $\mathcal{D}_w$ to embed watermarks while preserving pathology by aligning its output with the original VAE decoder \(\mathcal{D}_o\), using a pathology localization map $A_p^\text{loc}$ in the loss function to regulate watermark placement.
  } 
   \label{fig:overview}
\end{figure}

As illustrated in Fig.~\ref{fig:overview}, we propose MedSign, a deep watermarking framework for text-driven medical image synthesis, ensuring high-fidelity preservation of pathological features. Our approach consists of (1) pre-training a watermark extractor (Sec.~\ref{method:pretrain}) and (2) fine-tuning the LDM decoder to integrate watermarks while minimizing interference with diagnostically significant regions (Sec.~\ref{method:decoder}).

\subsection{Pre-training the watermark extractor}
\label{method:pretrain}
We train the watermark extractor using the HiDDeN architecture~\cite{zhu2018hidden} as illustrated in Fig.~\ref{fig:overview}(a). 
This architecture consists of: (1) a watermark encoder \(\mathcal{E}_m\) embedding a \(k\)-bit message into an image, (2) a transformation layer, and (3) a watermark extractor \(\mathcal{D}_m\) that retrieves the binary message. 
The watermark encoder \(\mathcal{E}_m\) embeds a \(k\)-bit message \(m\in\{0,1\}^k\) into an image \(\mathcal{I}\) to produce a watermarked image \(\tilde{\mathcal{{I}}}\). After applying image transformation \(T\) (e.g., cropping, rotation) to \(\tilde{\mathcal{{I}}}\), extractor \(\mathcal{D}_m\) recovers the message, formally represented as $\tilde{m} = \mathcal{D}_m(T(\tilde{\mathcal{{I}}}))$.
The model is trained by minimizing the binary cross-entropy loss $\mathcal{L}_{msg} = -\sum_{i=1}^{k} \left[m_i \log \sigma(\tilde{m}_i) + (1-m_i) \log (1-\sigma(\tilde{m}_i))\right]$.
Similar to \cite{fernandez2023stable}, image quality is not optimized (since \(\mathcal{E}_m\) is discarded); instead, PCA whitening is applied after training to remove bias and decorrelate outputs.  
Feature vectors from \( \mathcal{D}_m \) are centered using \( \mu = \mathbb{E}[\mathcal{D}_m] \), and the covariance matrix \( \Sigma = U\Lambda U^T \) is eigendecomposed. The whitening transform \( \Lambda^{-1/2} U^T \) is then applied, with the resulting bias \(-\Lambda^{-1/2} U^T \mu\) and weight \(\Lambda^{-1/2} U^T\) appended as a linear layer.

\subsection{Fine-tuning the LDM decoder}
\label{method:decoder}
We fine-tune the LDM's VAE decoder $\mathcal{D}_w$ to integrate watermarking directly into image generating process as shown in Fig.~\ref{fig:overview}(b). To preserve diagnostically critical regions, we use a pathology localization map from cross-attention in the diffusion process. This map guides the loss function, enabling adaptive watermarking while maintaining image quality.

\subsubsection{Pathology Localization Map Generation.}
\label{method:map}
To localize pathologically significant regions, we leverage cross-attention mechanisms~\cite{rombach2022high,ronneberger2015u} that connect the diffusion denoising U-Net’s spatial layout with medical text tokens.

\begin{figure}[t]
  \centering
  \includegraphics[width=\linewidth]{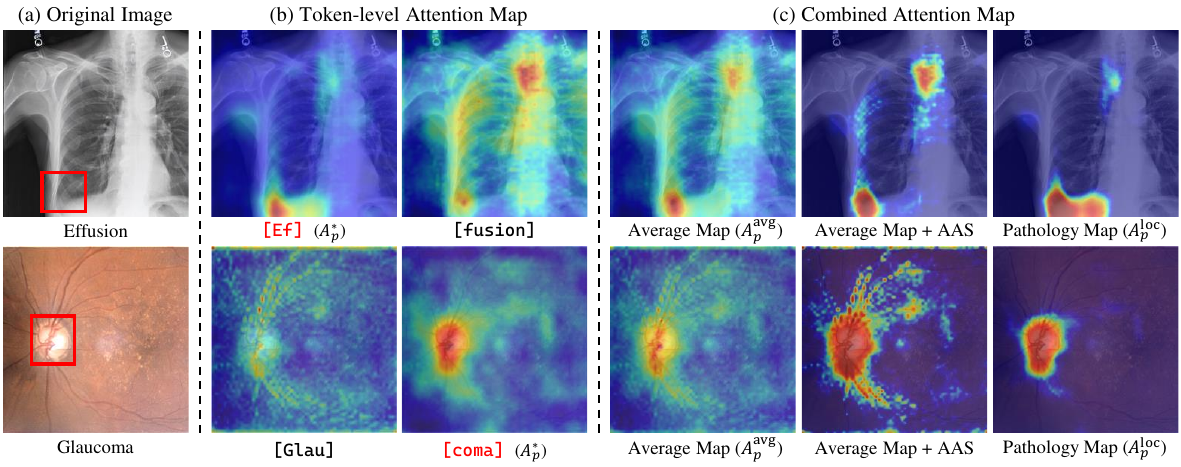}
  \caption{
  Visualization of token-level and combined attention maps with pathology localization map.
(a) Original CXR images with pathology regions in red boxes.
(b) Token-level attention maps: individual maps for tokens composing pathology-related words. 
The red-highlighted token's attention map is the one selected through CAS.
  (c) Combined attention maps: (left) the average attention map of all tokens, (middle) its refinement using AAS, and 
  (right) the final pathology localization map.
  } 
  \label{fig:attention}
\end{figure}

\paragraph{Cross-Attention for Pathology Localization.} 
We build on \cite{Marcos-Manchon_2024_CVPR} to aggregate attention maps from diffusion models, highlighting pathological regions in medical images. 
Specifically, a U-Net-based~\cite{ronneberger2015u} denoising network~\cite{rombach2022high} \(\epsilon_{\theta}(\mathbf{x}, t; \mathbf{y})\) maps noisy image features \(\mathbf{x}_t \in \mathbb{R}^{h\times w}\) at step \(t\) to a denoised output, conditioned on an input medical report \(\mathbf{y}\).  
Here, \(\mathbf{y}\) comprises medical words, each word $p$ consisting of $n$ tokens as \(p=\{s_1, s_2, \dots, s_n\} \). 
To incorporate textual guidance, cross-attention~\cite{rombach2022high,vaswani2017attention} injects token-level semantic embeddings from the medical report into the denoising network. This U-Net-based denoising network progressively refines the noisy input \(\mathbf{x}_t\), generating token-specific attention maps \(\mathbf{F}_s\), which correlate each token’s semantic content with its spatial layout (see Fig.~\ref{fig:attention}(b)).  
Then, each \(\mathbf{F}_s\) is upsampled to \(\widetilde{\mathbf{F}}_s\) to match the image resolution \((h, w)\), and the token-level maps are then aggregated across layers, attention heads, and time steps as $A_p^\text{avg}(x,y) = \frac{1}{n} \sum_{i=1}^{n} \sum_{l, h, t} \widetilde{\mathbf{F}}_{s_i}^{(l, h, t)}(x,y)$,
where \(\widetilde{\mathbf{F}}_{s_i}^{(l, h, t)}\) is the upsampled attention map for token \( s_i \) at layer \( l \), attention head \( h \), and time step \( t \). This aggregation captures the collective spatial influence of all tokens in \( p \), aligning attention with pathological regions in the medical image.  

\paragraph{Density Clustering-Driven Attention Selection.} 
While the aggregated map \( A_p^\text{avg} \) captures pathological regions, certain tokens within \( p \) can produce noise, shifting attention to irrelevant areas (Fig.~\ref{fig:attention}(c) left), which may prevent watermarking in permissible regions and degrade watermarking performance.
To address this, we adopt a density clustering-driven~\cite{ester1996density} attention selection mechanism (CAS) that suppresses dispersed attention signals and enhances pathological localization.
Given a medical word \( p \) consisting of tokens \( s_1, \dots, s_n \), we seek the most precise attention map by assuming that well-defined maps exhibit dense, localized distributions.  
Thus, for each token \( s_i \), we define a clustering function \( \mathcal{C} \) that estimates the number of dense clusters \( k_i \) in the spatial attention distribution of \( \widetilde{\mathbf{F}}_{s_i} \) using a density-based clustering algorithm, formulated as \( k_i = \mathcal{C}(\widetilde{\mathbf{F}}_{s_i}) \).  
To select the most localized attention map, we identify the token whose attention distribution exhibits the minimum cluster count as $s^* = \arg\min_{s_i} k_i$.
Hence, the final localized attention map for \( p \)  is defined as $A_p^*(x, y) = \widetilde{\mathbf{F}}_{s^*}(x, y)$.  
By selecting the attention map with the most compact distribution, this approach effectively mitigates noise and ensures precise pathological localization.

\paragraph{Adaptive Attention Scaling.}  
To improve the clarity of localization boundaries, we apply an Adaptive Attention Scaling (AAS) mechanism that refines the attention distribution. 
We first normalize the selected attention map \( A_p^* \) using z-score transformation, followed by smooth sigmoid scaling and global min-max normalization. 
To emphasize salient regions, we define an adaptive threshold \(\tau\) as the 0.7 quantile of \( A_p^* \), a value empirically determined to provide a favorable trade-off between watermark robustness and the preservation of critical diagnostic information. We then define our final pathology localization map as
\begin{equation}
    A_p^{\text{loc}}(x,y) \;=\; \sigma\,\bigl(A_p^*(x,y) \;-\;\tau\bigr),
    \label{eq1}
\end{equation}
where \(\sigma(\cdot)\) is the sigmoid function. As shown in Fig.~\ref{fig:attention}(c), this process enhances the contrast between pathological regions and the background, leading to sharper and more well-defined localization boundaries.

\subsubsection{Pathology-Aware Adaptive Loss.}  
\label{method:loss}  
We train the LDM decoder \( \mathcal{D}_w \) with a loss function comprising image perceptual loss (\(\mathcal{L}_{\text{img}}\)), message loss (\(\mathcal{L}_{\text{msg}}\)), total variation regularization (\(\mathcal{L}_{\text{tv}}\)), and pathology preservation loss (\(\mathcal{L}_{\text{pre}}\)). 

To preserve perceptual consistency, we use Watson-VGG perceptual loss~\cite{czolbe2020loss} for \(\mathcal{L}_{\text{img}}\), which minimizes visual distortions while keeping the watermark imperceptible.  
\(\mathcal{L}_{\text{msg}}\) is computed via binary cross-entropy between the ground-truth message \( m \) and the predicted message \( \hat{m} \), ensuring robust message recovery.  
To achieve this, we use the watermark extractor \(\mathcal{D}_m\) trained in Sec.~\ref{method:pretrain} by freezing its parameters during training.  
To encourage smooth watermark blending and suppress high-frequency artifacts, we incorporate \(\mathcal{L}_{\text{tv}}\), which penalizes sharp watermark variations.  
Meanwhile, the pathology preservation loss (\(\mathcal{L}_{\text{pre}}\)) prevents watermark insertion in critical regions by leveraging the pathology localization map \( A_p^{\text{loc}} \), preserving important areas while allowing embedding in less critical regions.  
The total objective \(\mathcal{L}_{\text{total}}\) integrates these components as  
\begin{equation}
    \footnotesize
    \begin{aligned}
    \mathcal{L}_{\text{total}} =&~ 
    \lambda_{\text{img}} \underbrace{\sum_{l} \delta_l \left\| \Bigl( \phi_l(\mathcal{I}_w) - \phi_l(\mathcal{I}_o) \Bigr) \right\|_2^2}_{\mathcal{L}_{{\text{img}}}}  
    ~+~ \lambda_{\text{msg}} \underbrace{\sum_{i=1}^{k} \text{BCE}(m_i, \hat{m}_i)}_{\mathcal{L}_{{\text{msg}}}} \\
    +&~ \lambda_{\text{tv}} \underbrace{\text{TV}(|\mathcal{I}_w - \mathcal{I}_o|)}_{\mathcal{L}_{\text{tv}}}
     ~+~ \lambda_{\text{pre}} \underbrace{\mathbb{E} \left[ \| (\mathcal{I}_w - \mathcal{I}_o) \odot A_p^{\text{loc}} \|_2^2 \right]}_{\mathcal{L}_{\text{pre}}},
    \end{aligned}
\end{equation} 
where the watermarked image is defined as \( \mathcal{I}_w = \mathcal{D}_w(\mathbf{x}_0) \), while the original (non-watermarked) image \( \mathcal{I}_o \) is decoded from the frozen original LDM decoder \( \mathcal{D}_o \) as \( \mathcal{I}_o = \mathcal{D}_o(\mathbf{x}_0) \). The VGG layer-\( l \) features, denoted as \( \phi_l(\cdot) \), are weighted by \( \delta_l \) to guide perceptual consistency.  

%% file: tex/experiments.tex
\section{Experiments}
\begin{figure}[t]
  \centering
    \includegraphics[width=\linewidth]{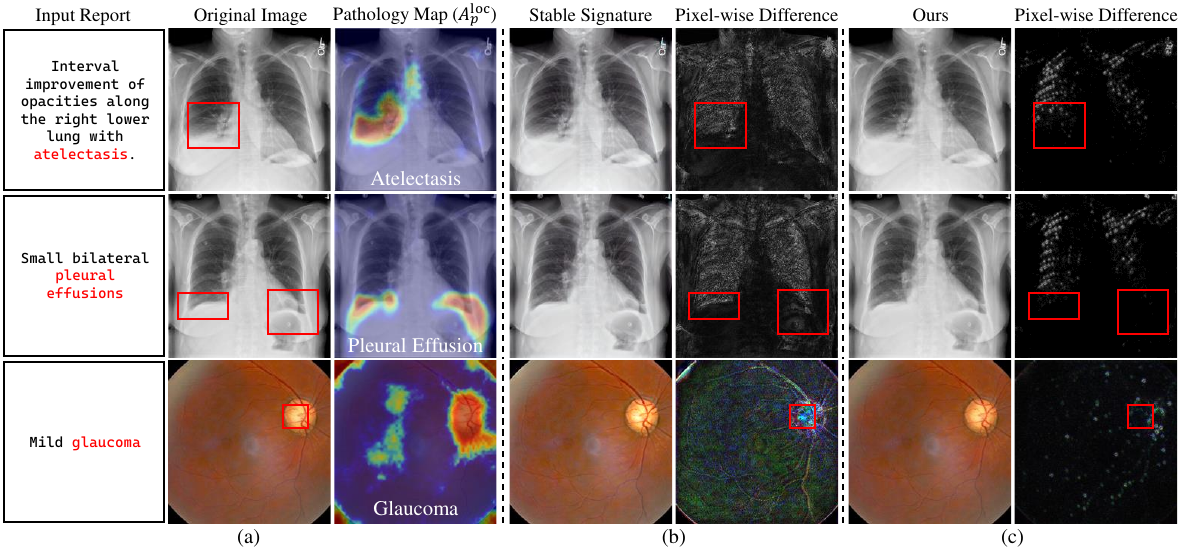}  
  \caption{
Qualitative comparison of Stable Signature~\cite{fernandez2023stable} and ours on MIMIC-CXR-JPG~\cite{johnson2019mimic} and OIA-ODIR~\cite{li2021benchmark}.  
(a): Original image with pathology localization map.  
(b) \& (c): Watermarked image with pixel-wise difference maps for Stable Signature and ours, respectively.  
Unlike Stable Signature, which embeds watermarks in critical regions (red box), our method preserves diagnostic areas.
}
  \label{fig:qualitative}
\end{figure}
\noindent
\textbf{Datasets.}
We use the impression section of reports from MIMIC-CXR-JPG~\cite{johnson2019mimic} for CXR images. For training, we randomly select 6,000 unique samples from the official train split after removing duplicates. For testing, we use 1,903 unique reports from the test set, also excluding duplicates.
For fundus data, we use unique reports from the OIA-ODIR dataset~\cite{li2021benchmark}. As images are generated from 290 distinct reports with different random seeds, no predefined split exists. Thus, we randomly assign 3,361 samples for training and 2,000 for testing.

\noindent
\textbf{Implementation Details.}
For diffusion model initialization, we use pretrained weights from RoentGen~\cite{Bluethgen2024} for CXR images and DERETFound~\cite{DERETFound} for fundus images.
For density-based clustering, we employ DBSCAN~\cite{ester1996density} with default settings.
The hyperparameters are set as follows: for CXR images, \( \lambda_{\text{img}} =2.1 \), \( \lambda_{\text{msg}} =7.5 \), \( \lambda_{{tv}} ={3.0} \) and \( \lambda_{\text{pre}} =975.0 \); and for fundus images, \( \lambda_{\text{img}} =1.9 \), \( \lambda_{\text{msg}} =6.0 \), \( \lambda_{\text{tv}} =10.0 \) and \( \lambda_{\text{pre}} =800.0 \), ensuring a balanced scale across them.
Training runs for 2 epochs using AdamW with a \( 5 \times 10^{-4} \) learning rate.

\subsection{Evaluation Results}
\subsubsection{Qualitative Analysis.}
We qualitatively compare our watermarked images with the in-generation watermarking method~\cite{fernandez2023stable} in Fig.~\ref{fig:qualitative}. 
The pixel-wise difference map, computed as \( | \mathcal{I}_w - \mathcal{I}_o | \times 10 \), highlights watermarking effects.  
(1) \textit{Pathology Localization Map}: Our method reduces misplaced alterations and artifacts by guiding watermark placement away from clinically significant areas using \( A_p^{\text{loc}} \).  
(2) \textit{Preservation of Critical Regions}: Integrating \( A_p^{\text{loc}} \) into training enhances watermark embedding in non-critical areas while preserving diagnostic integrity. Compared to the baseline, our approach minimizes interference in pathological regions (see red box in Fig.~\ref{fig:qualitative}(b) and (c)), ensuring minimal clinical impact.

\begin{table}[t]
\centering
\caption{Quantitative evaluation of watermarking performance on the MIMIC-CXR-JPG dataset~\cite{johnson2019mimic} and OIA-ODIR dataset~\cite{li2021benchmark}. 
Evaluated attack scenarios include None (no attack), Brt (brightness $\times2$), Crp (50\% center crop), JPG (JPEG compression 50\%), Rot (25° rotation), and Res (70\% scaling).
Method with \(\dagger\) uses 32-bit message encoding, while others use 48-bit.
We highlight the \textbf{best} and \underline{second-best} results.
}
\label{table:comparison}
\footnotesize
\setlength{\tabcolsep}{3pt}
\begin{tabular}{llcc*{6}{c}c}
\toprule[1.2pt]
\multirow{2.4}{*}{\textbf{}} & \multirow{2.4}{*}{\textbf{Method}} & \multirow{2.4}{*}{PSNR} & \multirow{2.4}{*}{SSIM} & \multicolumn{7}{c}{{Bit Accuracy}} \\
\cmidrule(lr){5-11}
& & & & None & Brt & Crp & JPG & Rot & Res  & {Avg} \\
\midrule
\rowcolor{gray!10}
\multicolumn{11}{c}{{{MIMIC-CXR-JPG}}} \\
\multirow{3}{*}{\makecell{Post\\Gen}} & DCT-DWT~\cite{al2007combined} & {35.17} & {0.89} & \textbf{1.00} & 0.51 & 0.50 & 0.52 & 0.51 & 0.53 & 0.59 \\
& SSL Watermark~\cite{fernandez2022watermarking} & \underline{36.20} & 0.89 & 0.93 & 0.66 & 0.84 & \underline{0.71} & \underline{0.85} & 0.87 & 0.81 \\
& WAM$^\dagger$~\cite{sander2025watermark} & 35.12 & \underline{0.93} & \textbf{1.00} & \textbf{1.00} & \underline{0.99} & 0.57 & 0.46 & \textbf{1.00} & 0.83 \\
\midrule
\multirow{2}{*}{\makecell{In\\Gen}} & Stable Signature~\cite{fernandez2023stable} & 33.24 & 0.92 & \underline{0.99} & \underline{0.99} & \underline{0.99} & 0.70 & 0.71 & 0.95 & \underline{0.89} \\
 & \cellcolor{blue!3}{Ours (MedSign)} & \cellcolor{blue!3}{\textbf{39.71}} & \cellcolor{blue!3}{\textbf{0.96}} & \cellcolor{blue!3}{\textbf{1.00}} & \cellcolor{blue!3}{\textbf{1.00}} & \cellcolor{blue!3}{\textbf{1.00}} & \cellcolor{blue!3}{\textbf{0.74}} & \cellcolor{blue!3}{\textbf{0.96}} & \cellcolor{blue!3}{\underline{0.97}} & \cellcolor{blue!3}{\textbf{0.95}} \\
\midrule
\rowcolor{gray!10}
\multicolumn{11}{c}{{{OIA-ODIR}}} \\
\multirow{3}{*}{\makecell{Post\\Gen}} & DCT-DWT~\cite{al2007combined} & 35.64 & \underline{0.90} & 0.81 & 0.49 & 0.49 & 0.51 & 0.50 & 0.80 & 0.60 \\
& SSL Watermark~\cite{fernandez2022watermarking} & \underline{36.37} & 0.89 & 0.92 & 0.84 & 0.84 & 0.71 & \underline{0.86} & 0.89 & 0.84 \\
& WAM$^\dagger$~\cite{sander2025watermark} & 31.35 & \textbf{0.94} & \textbf{1.00} & \textbf{1.00} & 0.79 & \underline{0.72} & 0.45 & \textbf{1.00} & 0.85 \\
\midrule
\multirow{2}{*}{\makecell{In\\Gen}} & Stable Signature~\cite{fernandez2023stable} & 30.20 & 0.79 & \underline{0.97} & \underline{0.97} & \underline{0.95} & 0.71 & 0.66 & 0.91 & \underline{0.86} \\
 & \cellcolor{blue!3}{Ours (MedSign)} & \cellcolor{blue!3}{\textbf{40.81}} & \cellcolor{blue!3}{\textbf{0.94}} & \cellcolor{blue!3}{\textbf{1.00}} & \cellcolor{blue!3}{0.95} & \cellcolor{blue!3}{\textbf{0.99}} & \cellcolor{blue!3}{\textbf{0.78}} & \cellcolor{blue!3}{\textbf{0.95}} & \cellcolor{blue!3}{\underline{0.97}} & \cellcolor{blue!3}{\textbf{0.94}}\\
\bottomrule[1.2pt]
\end{tabular}
\end{table}

\noindent\textbf{Quantitative Evaluation on Watermarking Performance.}
Table~\ref{table:comparison} evaluates watermarking performance on MIMIC-CXR-JPG~\cite{johnson2019mimic} and OIA-ODIR~\cite{li2021benchmark}, considering image quality (PSNR, SSIM) and watermarking robustness (Bit Accuracy) under perturbations.  
Methods are classified into \textit{Post-Generation Watermarking}, where watermarks are added after image generation, and \textit{In-Generation Watermarking}, where they are embedded during synthesis for direct comparison with ours.    
DCT-DWT~\cite{al2007combined}, used in Stable Diffusion~\cite{rombach2022high}, ensures perceptual fidelity but lacks robustness. SSL Watermark~\cite{fernandez2022watermarking} and WAM~\cite{sander2025watermark} enhance robustness but remain vulnerable to specific attacks, while Stable Signature~\cite{fernandez2023stable} balances detection at the cost of image fidelity.  
In contrast, MedSign ensures high-fidelity representation and resists attacks across CXR and fundus imaging, ensuring robustness across organs and modalities in medical watermarking.

\noindent\textbf{Quantitative Evaluation on Diagnostic Impact.}
To assess the diagnostic impact, Table~\ref{table:classification}(a) reports absolute differences in confidence scores (\%p) between watermarked and original images  (i.e., $|P_{\text{watermarked}}-P_{\text{original}}|\times100$) across six pulmonary diseases. Scores are obtained using a state-of-the-art CXR classifier~\cite{Cohen2022xrv}.  
A non-medical-specific watermarking model~\cite{fernandez2023stable} introduces an average confidence difference of 5.08, reaching 13.18 for Pneumonia, which could misrepresent clinical findings. In contrast, ours achieves a significantly lower average difference, minimizing diagnostic impact while preserving medical integrity.

\begin{table}[t]
\centering
\caption{ 
Comparative analysis on MIMIC-CXR-JPG~\cite{johnson2019mimic}.
(a) Diagnostic confidence: Absolute confidence score differences for six diseases. 
(b) Robustness and image quality: Average bit accuracy and PSNR, assessing the general performance of different models. 
}
\label{table:classification}
\footnotesize
\setlength{\tabcolsep}{1pt}
\begin{tabular}{l*{7}{c}ccc}
\toprule[1.2pt]
\multirow{2.4}{*}{\textbf{Method}} & \multicolumn{7}{c}{(a)} & & \multicolumn{2}{c}{(b)} \\
\cmidrule(lr){2-8} \cmidrule(lr){10-11}
& Pna. & Pmtx. & Atel. & Ede. & Cmgl. & Effu. & Avg $(\downarrow)$ & & AvgAcc $(\uparrow)$  &  PSNR $(\uparrow)$\\
\midrule
Stable Signature~\cite{fernandez2023stable} & 13.18 & 2.76 & 1.99 & 7.87 & 2.99 & 1.70 & 5.08 & & 0.89 & 33.24 \\
\rowcolor{blue!3}
Ours (MedSign) & \textbf{1.83} & \textbf{0.36} & \textbf{0.33} & \textbf{0.85} & \textbf{0.02} & \textbf{0.15} & \textbf{0.59} & & \textbf{0.95} & \textbf{39.71} \\
\midrule
(1) Ours w/o AAS   & 1.68 & 0.45 & 0.23 & 0.73 & 0.06 & 0.13 & 0.55 & & 0.94 & 38.65 \\
(2) Ours w/o CAS   & 1.75 & 0.38 & 0.25 & 0.75 & 0.06 & 0.16 & 0.56 & & 0.92 & 39.51 \\
(3) Ours w/o \( \mathcal{L}_{\text{tv}} \)    & 1.79 & 0.47 & 0.37 & 0.84 & 0.11 & 0.14 & 0.62 & & 0.95 & 37.75 \\
(4) Ours w/o \( \mathcal{L}_{\text{pre}} \)  & 4.20 & 1.91 & 2.27 & 2.31 & 1.23 & 1.89 & 2.30 & & 0.94 & 35.02 \\
\bottomrule[1.2pt]
\end{tabular}
\end{table}

\subsection{Ablation Analysis}
\noindent \textbf{(1) \& (2): Effect of AAS and CAS.}
We analyze the impact of AAS and CAS by removing each from MedSign in Table~\ref{table:classification}. Here, (2) refers to the use of \( A_p^\text{avg} \) instead of $A_p^*$ in Eq.~\eqref{eq1}. 
Diagnostic confidence remains stable since the removal enforces watermarking within a narrower region than our proposed pathology localization map.
However, this constraint reduces average bit accuracy, highlighting the trade-off between diagnostic integrity and watermark robustness.

\noindent \textbf{(3): Effect of \( \mathcal{L}_{\text{tv}} \).}
Total variation regularization \( \mathcal{L}_{\text{tv}} \) enhances the smoothness of watermarking, preserving image fidelity while maintaining diagnostic consistency as shown in Table~\ref{table:classification}(3). Its removal reduces PSNR and introduces visual artifacts, though diagnostic confidence remains largely unaffected.

\noindent \textbf{(4): Effect of \( \mathcal{L}_{\text{pre}} \).}
As shown in Table~\ref{table:classification}(4), removing pathology preservation loss \( \mathcal{L}_{\text{pre}} \) leads to a significant drop in PSNR and increased alterations in diagnostic confidence, while average bit accuracy remains nearly unchanged. Without \( \mathcal{L}_{\text{pre}} \), watermarking is applied globally, preserving bit accuracy but compromising diagnostic integrity and degrading overall image quality. This suggests that \( \mathcal{L}_{\text{pre}} \) is essential for restricting watermarking to non-critical regions, maintaining diagnostic integrity, and preserving clinically significant features.

%% file: tex/conclusion.tex
\section{Conclusion and Discussion}
We present a pathology-aware watermarking framework for text-to-medical image generation, ensuring robust embedding while preserving diagnostically critical regions. 
By leveraging a carefully tailored cross-attention map, our method precisely localizes pathologies and regulates watermark placement to minimize interference. 
Experiments show that our approach balances image fidelity and watermark resilience, surpassing existing methods in robustness while minimizing clinical interpretation.
This study highlights the potential of pathology-aware watermarking to authenticate text-driven AI-generated medical images without compromising diagnostic utility. 
Future work includes extending our approach to other modalities, such as CT and MRI, for broader medical applicability.

%% file: Paper-0428.bbl
\begin{thebibliography}{10}
\providecommand{\url}[1]{\texttt{#1}}
\providecommand{\urlprefix}{URL }
\providecommand{\doi}[1]{https://doi.org/#1}

\bibitem{al2007combined}
Al-Haj, A.: Combined dwt-dct digital image watermarking. Journal of computer science  (2007)

\bibitem{Bluethgen2024}
Bluethgen, C., Chambon, P., Delbrouck, J.B., van~der Sluijs, R., Po{\l}acin, M., Zambrano~Chaves, J.M., Abraham, T.M., Purohit, S., Langlotz, C.P., Chaudhari, A.S.: A vision--language foundation model for the generation of realistic chest x-ray images. Nature Biomedical Engineering  (2024)

\bibitem{ci2024ringid}
Ci, H., Yang, P., Song, Y., Shou, M.Z.: Ringid: Rethinking tree-ring watermarking for enhanced multi-key identification. In: ECCV (2024)

\bibitem{Cohen2022xrv}
Cohen, J.P., Viviano, J.D., Bertin, P., Morrison, P., Torabian, P., Guarrera, M., Lungren, M.P., Chaudhari, A., Brooks, R., Hashir, M., Bertrand, H.: {TorchXRayVision: A library of chest X-ray datasets and models}. In: Medical Imaging with Deep Learning (2022)

\bibitem{czolbe2020loss}
Czolbe, S., Krause, O., Cox, I., Igel, C.: A loss function for generative neural networks based on watson’s perceptual model. In: NeurIPS (2020)

\bibitem{ester1996density}
Ester, M., Kriegel, H.P., Sander, J., Xu, X., et~al.: A density-based algorithm for discovering clusters in large spatial databases with noise. In: KDD (1996)

\bibitem{EU_AI_Act_2024}
{European Parliament and Council}: {Regulation (EU) 2024/1689 of the European Parliament and of the Council on Artificial Intelligence} (2024), \url{https://eur-lex.europa.eu/legal-content/EN/TXT/?uri=CELEX:32024R1689}

\bibitem{fernandez2023stable}
Fernandez, P., Couairon, G., J{\'e}gou, H., Douze, M., Furon, T.: The stable signature: Rooting watermarks in latent diffusion models. ICCV  (2023)

\bibitem{fernandez2022watermarking}
Fernandez, P., Sablayrolles, A., Furon, T., J{\'e}gou, H., Douze, M.: Watermarking images in self-supervised latent spaces. In: International Conference on Acoustics, Speech, and Signal Processing (2022)

\bibitem{2024cxrl}
Han, W., Kim, C., Ju, D., Shim, Y., Hwang, S.J.: Advancing text-driven chest x-ray generation with policy-based reinforcement learning. In: MICCAI (2024)

\bibitem{han2025spatial}
Han, W., Lee, Y., Kim, C., Park, K., Hwang, S.J.: Spatial transport optimization by repositioning attention map for training-free text-to-image synthesis. In: CVPR (2025)

\bibitem{ho2020denoising}
Ho, J., Jain, A., Abbeel, P.: Denoising diffusion probabilistic models. NeurIPS  (2020)

\bibitem{jang2024waterf}
Jang, Y., Lee, D.I., Jang, M., Kim, J.W., Yang, F., Kim, S.: Waterf: Robust watermarks in radiance fields for protection of copyrights. In: CVPR (2024)

\bibitem{johnson2019mimic}
Johnson, A.E., Pollard, T.J., Greenbaum, N.R., Lungren, M.P., Deng, C.y., Peng, Y., Lu, Z., Mark, R.G., Berkowitz, S.J., Horng, S.: Mimic-cxr-jpg, a large publicly available database of labeled chest radiographs. arXiv preprint arXiv:1901.07042  (2019)

\bibitem{DERETFound}
Jonlysun: Deretfound: Deep embedded representation for fundus image analysis. \url{https://github.com/Jonlysun/DERETFound} (2023)

\bibitem{lee2024llmcxr}
Lee, S., Kim, W.J., Chang, J., Ye, J.C.: {LLM}-{CXR}: Instruction-finetuned {LLM} for {CXR} image understanding and generation. In: ICLR (2024)

\bibitem{li2021benchmark}
Li, N., Li, T., Hu, C., Wang, K., Kang, H.: A benchmark of ocular disease intelligent recognition: One shot for multi-disease detection. In: Benchmarking, Measuring, and Optimizing: Third BenchCouncil International Symposium, Bench 2020, Virtual Event, November 15--16, 2020, Revised Selected Papers 3. Springer (2021)

\bibitem{Marcos-Manchon_2024_CVPR}
Marcos-Manch\'on, P., Alcover-Couso, R., SanMiguel, J.C., Mart{\'\i}nez, J.M.: Open-vocabulary attention maps with token optimization for semantic segmentation in diffusion models. In: CVPR (2024)

\bibitem{medghalchi2024meddap}
Medghalchi, Y., Zakariaei, N., Rahmim, A., Hacihaliloglu, I.: Meddap: Medical dataset enhancement via diversified augmentation pipeline. arXiv preprint arXiv:2403.16335  (2024)

\bibitem{mirsky2019ct}
Mirsky, Y., Mahler, T., Shelef, I., Elovici, Y.: $\{$CT-GAN$\}$: Malicious tampering of 3d medical imagery using deep learning. In: 28th USENIX Security Symposium (USENIX Security 19) (2019)

\bibitem{Korea_Law_2024}
{National Assembly of the Republic of Korea}: {Act No. 20676} (2024), \url{https://www.law.go.kr/LSW/lsInfoP.do?lsiSeq=268543&viewCls=lsRvsDocInfoR#}

\bibitem{glide}
Nichol, A., Dhariwal, P., Ramesh, A., Shyam, P., Mishkin, P., McGrew, B., Sutskever, I., Chen, M.: Glide: Towards photorealistic image generation and editing with text-guided diffusion models. arXiv preprint arXiv:2112.10741  (2021)

\bibitem{ramesh2022hierarchical}
Ramesh, A., Dhariwal, P., Nichol, A., Chu, C., Chen, M.: Hierarchical text-conditional image generation with clip latents. arXiv preprint arXiv:2204.06125  (2022)

\bibitem{rombach2022high}
Rombach, R., Blattmann, A., Lorenz, D., Esser, P., Ommer, B.: High-resolution image synthesis with latent diffusion models. In: CVPR (2022)

\bibitem{ronneberger2015u}
Ronneberger, O., Fischer, P., Brox, T.: U-net: Convolutional networks for biomedical image segmentation. In: MICCAI (2015)

\bibitem{sander2025watermark}
Sander, T., Fernandez, P., Durmus, A., Furon, T., Douze, M.: Watermark anything with localized messages. In: ICLR (2025)

\bibitem{siracusano2023effective}
Siracusano, G., La~Corte, A., Nucera, A.G., Gaeta, M., Chiappini, M., Finocchio, G.: Effective processing pipeline pace 2.0 for enhancing chest x-ray contrast and diagnostic interpretability. Scientific Reports  (2023)

\bibitem{China_AI_Plan_2023}
{State Council of the People’s Republic of China}: {New Generation Artificial Intelligence Development Plan} (2023), \url{https://www.gov.cn/zhengce/content/202306/content_6884925.htm}

\bibitem{trabucco2024effective}
Trabucco, B., Doherty, K., Gurinas, M.A., Salakhutdinov, R.: Effective data augmentation with diffusion models. In: ICLR (2024)

\bibitem{vaswani2017attention}
Vaswani, A., Shazeer, N., Parmar, N., Uszkoreit, J., Jones, L., Gomez, A.N., Kaiser, L.u., Polosukhin, I.: Attention is all you need. In: NeurIPS (2017)

\bibitem{wen2023tree}
Wen, Y., Kirchenbauer, J., Geiping, J., Goldstein, T.: Tree-rings watermarks: Invisible fingerprints for diffusion images. NeurIPS  (2023)

\bibitem{Yellapragada_2024_WACV}
Yellapragada, S., Graikos, A., Prasanna, P., Kurc, T., Saltz, J., Samaras, D.: Pathldm: Text conditioned latent diffusion model for histopathology. In: WACV (2024)

\bibitem{zhang2024editguard}
Zhang, X., Li, R., Yu, J., Xu, Y., Li, W., Zhang, J.: Editguard: Versatile image watermarking for tamper localization and copyright protection. In: CVPR (2024)

\bibitem{zhu2018hidden}
Zhu, J., Kaplan, R., Johnson, J., Fei-Fei, L.: Hidden: Hiding data with deep networks. In: ECCV (2018)

\end{thebibliography}
